\documentclass[conference]{IEEEtran}
\IEEEoverridecommandlockouts
\usepackage{amsmath,amssymb,amsfonts}
\usepackage{algorithmic}
\usepackage{graphicx}
\usepackage{textcomp}
\usepackage{xcolor}
\usepackage{tabularx}
\usepackage{booktabs}
\usepackage{adjustbox}
\usepackage{nameref}
\usepackage{natbib}
\emergencystretch=\maxdimen
\def\BibTeX{{\rm B\kern-.05em{\sc i\kern-.025em b}\kern-.08em
    T\kern-.1667em\lower.7ex\hbox{E}\kern-.125emX}}
\begin{document}

\title{A Large Language Model-based Framework for Semi-Structured Tender Document Retrieval-Augmented Generation}

\author{\IEEEauthorblockN{Yilong Zhao}
\IEEEauthorblockA{\textit{School of Information Management} \\
\textit{Sun Yat-sen University}\\
Guangzhou, China \\
zhaoylong5@mail2.sysu.edu.cn}
\and
\IEEEauthorblockN{Daifeng Li*}
\IEEEauthorblockA{\textit{School of Information Management} \\
\textit{Sun Yat-sen University}\\
Guangzhou, China \\
lidaifeng@mail.sysu.edu.cn\\
*corresponding author}
}
\maketitle

\begin{abstract}
The drafting of documents in the procurement field has progressively become more complex and diverse, driven by the need to meet legal requirements, adapt to technological advancements, and address stakeholder demands. While large language models (LLMs) show potential in document generation, most LLMs lack specialized knowledge in procurement. To address this gap, we use retrieval-augmented techniques to achieve professional document generation, ensuring accuracy and relevance in procurement documentation.
\end{abstract}


\section{Introduction}
Tender documents are crucial in the procurement field, providing essential information and guidelines for acquiring goods, services, or works. The quality and accuracy of tender documents are vital for the success of the procurement. Traditional manual methods of preparing tender documents are time-consuming, error-prone, and inconsistent, posing challenges for non-professional procurement personnel. 

LLMs demonstrate significant potential in context understanding, with a multitude of recent studies concentrating on text generation across various fields and tasks. Nonetheless, the task of generating government procurement documents presents a higher degree of rigor compared to other endeavors. Such documents must conform to a certain structure and style, align precisely with the user's stipulated requirements, and not only comply with but also explicitly underscore the pertinent policies in accordance with the project's specifics. When it comes to the intricacies of detailed user requirements in procurement documentation, LLMs may fall short in domain-specific knowledge, thus lacking the capability to autonomously produce an original design.

To address this, we develop a framework leveraging natural language processing and artificial intelligence to automate tender document generation. Our main contributions to this work can be summarized as follows:
\begin{itemize}
    \item We introduce a LLM-based framework for the generation of semi-structured tender documents.
    \item We design a retrieval-augmented generation model, document indexing, retrieval based on the user requirement, document re-rank based on procurement lists.
     \item We use Word templates base on historical tender documents with smart tags to generate information. LLMs act as agents, interacting with users via smart tag instructions to retrieve information and generate content.
    \item Leveraging procurement project information and documents, we establish a knowledge database comprising a commodity classification knowledge base and a knowledge graph constructed with GraphRAG. Subsequently, we utilize this knowledge base to amend documents during the generation process.
\end{itemize}

\section{Related Work}

The utilization of LLMs in document generation has attracted significant interest, particularly in the field of public administration. Researchers have primarily focused on leveraging the advanced capabilities of LLMs for tasks such as document summarization and content generation within specific domains. These models have demonstrated their effectiveness in improving the efficiency and accuracy of document creation, as well as aiding in information extraction from large document collections. 

Recent research predominantly centers around leveraging LLMs to generate content within predefined thematic frameworks. This includes tasks like generating lecture abstracts, producing semi-structured documents \cite{musumeci2024llm}, lecture notes \cite{asthana23field}, and PPT \cite{christensen2023llm} generation. These tasks have fewer requirements for content and structure, making them well-suited for employing straightforward prompt-based approaches. Furthermore, there are document generation tasks that demand a high level of expertise and professionalism, such as the creation of legal documents. Lin et al. have proposed a prompt writing design leveraging large models and developed methods to evaluate contract terms specifically for legal document generation \cite{lin2024legal}. In order to protect information privacy, Lam et al. used a large number of unlabeled legal documents without Chinese word segmentation to fine-tune a LLM, in order to generate legal document drafts with local LLM \cite{lam2023applying}. Currently, it is common practice to optimize the prompt engineering and fine-tuning for document generation based on a certain framework structure and specific information. LLMs have demonstrated significant potential in document generation, leveraging their semantic understanding to enhance efficiency and accuracy in various document generation tasks. However, further research and exploration are still needed to address some challenging issues, such as structured content generation, compliance, and professional requirements.

Retrieval-augmented generation has gained attention in text generation tasks. Previous studies primarily use the training corpus or external datasets as retrieval sources, which can include external datasets \cite{zheng2021adaptive} or unsupervised data \cite{cai-etal-2021-neural}, enhance the model's knowledge and adaptability. Retrieval metrics in retrieval-augmented generation include sparse-vector retrieval (using methods like TF-IDF and BM25), dense-vector retrieval (leveraging pretrained language models for encoding text), and task-specific retrieval (learning a customized retrieval metric optimized for specific objectives). These metrics enhance the relevance and quality of retrieved examples to improve the final generation output. Retrieval metrics play a vital role in selecting relevant examples. These metrics can be classified into sparse-vector retrieval, which involves methods like TF-IDF and BM25 \cite{robertson2009probabilistic}, enabling efficient keyword matching using inverted indexes. On the other hand, dense-vector retrieval leverages pre-trained language models, such as those based on BERT, to encode text into low-dimensional dense vectors \cite{lee2019latent}. Retrieval-augmented text generation has emerged as a promising approach in computational linguistics. Its effectiveness and success rely on three key components: retrieval source, retrieval metric, and integration methods.

\section{Methods}
In our framework, the generation of tender documents typically involves a three-step process: 
\begin{itemize}
    \item First, when provided with a new procurement requirement like user basic information or procurement content, utilize the retrieval-augmentation module to find the most similar procurement document.
    \item Second, use a memory network to confirm the coherence among the retrieved document, policies, and new information, and subsequently make modifications to the document based on this verification.
    \item Third, leverage the procurement project knowledge base to further disassemble and modify the procurement content of procurement documents based on user procurement content.
\end{itemize}

Building upon this process, we execute document generation. The framework comprises three key components: the template retrieval module, the verification module with a memory network, and the modification module with the procurement knowledge base. The detailed structure is depicted in Figure \ref{10}.

\subsection{Template Retrieval Module}
In the realm of procurement professions, historical tender documents are gathered to form a document corpus $D = \{ d_i \}^{|D|}_{i=1}$. This corpus encapsulates essential details like the project name, purchaser unit, purchase content list, and more, organized as fields. These details are depicted as $F = \{ f_{1}, f_{2}, ..., f_{m} \}$, where $m$ represents the count of fields within a document. The template retrieval module typically collaborates with a retriever capable of fetching a document list $D_r = res(r)$ for purchaser to choose a template based on a purchase requirement $r$. This module consists of two stages: a retrieval stage that utilizes fields to acquire initial document lists and a re-ranking stage that utilizes lists of purchase content to refine the results.

\subsubsection{Retrieval Stage}
In first stage, we combine embedding-based retrieval and vocabulary-based retrieval. In the retrieval process, each requirement field conducts retrieval separately and obtain an aggregation of a document list $D_{terms}$ linked to every term from this field through index. To create the embedding index, we leverage a BERT-based encoder with a Contriever architecture. Given a document requirement field We get embedding index scores $d_score_e$ for each document through a faiss contriever. And for vocabulary-based search, we implement a multi-term indexing denoted as $index\_v$. Terms' importance varies; we utilize the term weight $weight_{t_i}$ to indicate the significance of the $i$-th term in the document collection and employ $frequency_d$ to denote the frequency of document occurrences for requirement terms. Subsequently, the vocabulary indexing score for each document can be calculated by averaging document frequency and the sum of term weights\. Then, we can derive the document index score for each document by summing all scores across all fields. The initial document rank list will be obtained based on the order of the document index scores.

\begin{equation}
D_{t_i}  = \begin{array}{ll}
index\_v[t_i] & \text{if } t_i \text{ exists in index\_v} 
\end{array}
\end{equation}
\begin{equation}
\small{
weight_{t_i} = \frac{\sum \left( len(D_{t}) + 1\right) }  {len(D_{t_i})} \mathbin{/}  {\sum_{k} \left( \frac{\sum \left( len(D_{t}) + 1\right)} { len(D_{t_k}) + 1 }\right)}
}
\end{equation}
\begin{equation}
frequency_d  = \frac{ |d \in \sum D_{t}| }{|t|}
\end{equation}
\begin{equation}
d\_score\_v =  Avg\left(\sum^{|t_i \in d|} weight_{t_i}, frequency_d \right)
\end{equation}
\begin{equation}
frequency_d  = \frac{ |d \in \sum D_{t}| }{|t|}
\end{equation}
\begin{equation}
\small{
d\_score =  \sum_{f_i}^{f_m} Avg\left( d\_score\_e + d\_score\_v \right)
}
\end{equation}
\subsubsection{Re-rank Stage}
One of the most crucial distinguishing features of procurement documents is the purchase lists, each with unique requirements and policy terms associated with different procurement items. When a purchaser submits a purchase list, the document lists obtained from the previous step will be re-prioritized according to the similarity between their historical purchase lists and the requirements of the current purchase list.
In this process, we analyze similarity on an individual item basis and then iterate through each historical list denoted as $h\_list$ to aggregate similarity scores by comparing the item in the current list $c\_list$ with each one in each $h\_list$. The similarity metrics employed include n-gram distance, edit distance, and embedding cosine distance. To mitigate significant differences in list lengths, when computing the overall score for $h\_list$ , we aggregate the similarity scores of each item in $c\_list$  and then adjust by adding the difference in the lengths of the two lists as a penalty term.
\begin{equation}
\label{item_dist}
\small{
item\_dist =   \frac{ embedding\_dist - n\_gram\_dist +  - edit\_dist }{3}
}
\end{equation}
\begin{equation}
\small{
list\_dist =  \frac{ \sum^{|item|} item\_dist_i\\
+ \alpha |len(h\_list) - len(c\_list)|} { len(c\_list)}
}
\end{equation}

The document of highest similarity score will be used as the template in next process.

\subsection{Template Filling Module with Smart Tags}
The detailed information provided by a new purchaser may deviate from the retrieved document. We utilize LLM to adjust the document based on the new purchaser's information. before storing tender documents in the document corpus We incorporate smart tags for procurement parameters to preprocess them. 

Our framework facilitates seamless customization of tender documents by utilizing a standardized style in Word templates. The public tender notice should include important information such as the procuring entity, project details (including bidder qualifications, tender document availability, and bid submission deadline), and contact person's information. Prior to generating the tender document, users must provide us with the project details. These variables empower users to reference and manipulate data within the template. We apply the information retrieval agent for the task one, as shown in Table \ref{User's Information Retrieval Agent Task}.

\begin{table}
    \caption{User's Requirement Parameter Retrieval Agent Task}
    \begin{adjustbox}{width=0.45\textwidth}
    \begin{tabular}{p{8cm}}
      \toprule
      Prompt Engineering for the User's Information Parameter Filling\\
      \midrule
      \textbf{Accumulated information:} $r$\\
      \midrule
      \textbf{Task description:} \newline
          I want you to fill this template $\{\{t\}\}$  \newline
      \hspace{2cm} What is the missing information?\\
      \midrule
      \textbf{Agent retrieval prompt:} \newline
      You are an assistant helping to search which information is missing. The available information is: \newline
      \textless \textbf{Accumulated information} \textgreater  \newline
      Your output must be like this: the missing information to satisfy the request is INFORMATION\_MISSING. \newline
      If you have the information write the token \text{[ALL\_INFO]}. Strictly respond with only the information that is missing. \newline
      \textless \textbf{Task description} \textgreater \\
      \bottomrule
    \end{tabular}
    \end{adjustbox}
    \label{User's Information Retrieval Agent Task}
\end{table}

The retrieval agent will interact with the user until all the necessary information is obtained. Then, we utilize the content generation agent with instructions provided in Prompt 1.

\begin{quote}
\textbf{Task: Smart Tags Filling} \\
\textbf{Prompt 1.} \textit{You are an assistant with the purpose of generating a document with the available information. You have the following information: \textbf{Accumulated information}. Please fill the template $t_j$.}
\label{Prompt 1}
\end{quote}

In such cases, we can utilize a Word plugin named as PageOffice or the Python package python-docx to modify the size, content, and style of both paragraphs and tables. 
 
\subsection{Modification Module with Procurement Knowledge Base}
Procurement item lists vary among projects. Some projects only provide procurement project names or purposes without specifying detailed item lists or parameters for each item. In such cases, historical procurement project information is utilized to identify closely related procurement items. We apply GraphRAG on the tender document corpus, and analyze the resulting knowledge graph of items from previous projects using the graph query language Neo4j.

\begin{figure}[h]
  \centering
  \includegraphics[width=0.5\textwidth]{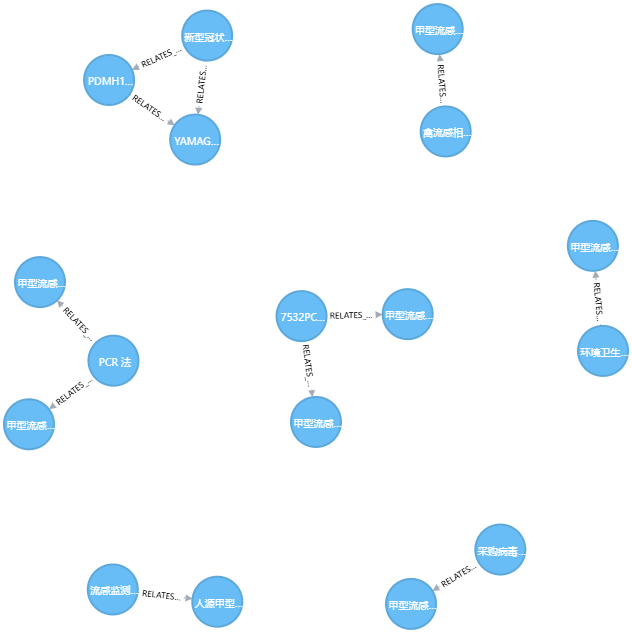} 
  \caption{Example of GraphRAG Query Result} %caption是图片的标题
  \label{3} %此处的label相当于图片的id，便于上下文的引用
\end{figure}

When the purchaser fails to provide a procurement list, we search the GraphRAG graph using Cypher language. For instance, we use Cypher query to obtain entities related to influenza A virus, the result graph is shown in Fig \ref{3}.
\begin{quote}
\textbf{Cypher Query:} \\
\text{MATCH (e:Entity)-[r]-(related)}
\text{where e.name contains "influenza A virus"}
\text{RETURN e,r,related}
\end{quote}

The utilization of an external product knowledge base enhances the efficiency of processing tender documents by categorizing items to align closely with procurement needs. This ensures relevance to user requirements and helps eliminate irrelevant goods. Once a product classification knowledge base is established, we distinguish procurement items from unrelated ones in historical documents based on the item embedding distance, with $\theta$ representing the threshold, as illustrated in Equation \ref{item_dist}.
\section{Experiment}

\subsection{Dataset}
We evaluated our framework using a dataset of 1406 tender documents containing procurement details, such as project name, project number, purchase item list, and more, from previous tender projects in the medical field. These documents were provided by a procurement group.

Based on the tender documents, we built a knowledge graph using GraphRAG with the DeepSeek LLM. The details of this graph are depicted in Figure \ref{graphrag}.

\begin{table}[h]
\centering
\caption{Node Label and Count in the Graph}
\label{graphrag}
\begin{tabular}{ccl} 
    \toprule
    Type & Count & Examples\\ 
    \midrule
    Entity	& 4536 & Microbial detection reagents, Test kit, ...\\
    Relationship & 3373 & is a kind of, is the procurement item for, ...\\
    \bottomrule
\end{tabular}
\end{table}

\subsection{Evaluation}
The tender documents are stored in an editable Word format, consisting primarily of paragraphs and tables. We evaluate the semantic accuracy of paragraph meanings and the content within tables based on the embedding similarity of the generated documents and the project document provided by the procurement team as gold standard.

To access each generated paragraph $p_i$, we iterate to calculate the similarity between $p_i$ and each paragraph from the gold standard $g\_s\_p$, aggregating the minimum distance scores as the similarity score of $p_i$.

\begin{equation}
\small{
score_{p_i} = \max \left( 1 - dist(p_i, g\_s\_p_j) \right),\quad j=0,1,...
}
\end{equation}
\begin{equation}
\small{
para\_score = \frac{1 - |len(p) - len(g\_s\_p)|}{MAX(len(p),len(g\_s\_p))} \times \sum_{i=0}^{|p|} \left( score_{p_i} \right) 
}
\end{equation}
For table evaluation, we follow a two-step process. Since different procurement content requires distinct parameters, we initially assess the similarity of the field name lists between table from generated document $t$ and table from gold standard $g\_s\_t$. Subsequently, building upon this comparison, we calculate the similarity of the table contents on a per-entry basis.
\begin{equation}
\small{
table\_score_{t_i} = Avg( \sum^{|field|} ( 1- dist{field}), list\_dist)
}
\end{equation}
\begin{equation}
\small{
table\_score = \frac{\sum_{i=0}^{|t|} table\_score_{t_i}} {len(t)}
}
\end{equation}
\begin{equation}
\small{
score = \frac{ len(p)\times para\_score + len(t) \times table\_score} {len(p) + len(t)}
}
\end{equation}

\subsection{Comparison with Baselines}
We randomly selected dozens of documents from the tender document repository to analyze and evaluate using our framework. The framework includes both the template retrieval module and the modification module, integrated with a procurement knowledge base, all utilizing a foundational base model. We have chosen ChatGLM-4 as the primary base model. In comparison, we are using ChatGLM-4 as the baseline alongside the incorporation of several modules that we have proposed separately. When using a basic large language model, the generated content may seem too arbitrary without a structured template like a bidding document. Additionally, there might be a lack of organization when it comes to processing and generating form data. While template-driven language models usually yield more coherent results, the style and format of the documents they produce can still deviate significantly from standard procurement files.

\begin{table}[h]
\centering
\caption{Performance on tender document dataset}
\label{1}
\begin{tabular}{lccc}
    \toprule
    Method & paragraph score & table score & score\\ 
    \midrule
    our framework & 78.31&     76.15  &   77.74\\
    ChatGLM-4    &  12.55 &       0& 12.55\\
    ChatGLM-4 with retrieval module  & 38.27 &      15.23 &    29.42 \\
    \bottomrule
\end{tabular}
\end{table}

\subsection{Ablations}
To demonstrate the impact of the proposed mechanisms, we dismantled elements of our framework to assess their individual contributions to the overall performance. By isolating and analyzing the Template Retrieval and Template Filling modules separately, we were able to gain insights into their specific effects on the paragraph and table scores. The results underscored the importance of each module in enhancing the overall score and highlighted the synergistic relationship between them within our framework. When we randomly select a template from the tender document corpus to assess the impact of removing the template retrieval module, we observe a significant decrease in both paragraph and table scores by several percentage points. The template filling module plays a crucial role in aligning the retrieved template with the given new requirements, with its impact on the score being relatively more significant compared to the template retrieval module.

\begin{table}[h]
\centering
\caption{Performance when remove some components of our model. \emph{r.m.} strands for remove. \emph{r.p.} stands for replaced with.}
\label{tab:ablation}
\begin{tabular}{lccc}
\toprule
Ablation Models &   paragraph score & table score & score\\ 
\midrule
Our Framework  & 78.31&     76.15  &   77.74\\
\emph{r.m.} Template Retrieval module     &74.23  &71.44 &73.47\\
\emph{r.m.} Template Filling module       &77.86 &    75.19&     76.54\\

\bottomrule
\end{tabular}
\end{table}

\section{Conclusions}
Our study proposes a feasible framework for tender document generation, utilizing retrieval-augmented techniques and LLMs. The document generation process is mainly completed in two steps: selecting a template and then modifying it. During this process, multiple LLM-based agents interact with the user, verifying the necessary information for creating the tender document and generating content based on user requirements. In future work, we will focus on expanding the tender document corpus and exploring methods for handling complex table tasks.

% \bibliography{reference.bib}

\end{document}